%% file: main.tex
\documentclass[letterpaper]{article}
\usepackage{aaai}
\nocopyright
\usepackage{times}
\usepackage{helvet}
\usepackage{courier}
\usepackage{amsmath}
\usepackage{amssymb}
\usepackage{booktabs}
\usepackage{multirow}
\usepackage{graphicx}
\usepackage{multicol}
\usepackage{stfloats}
\usepackage{placeins}
\usepackage{balance}
\graphicspath{{figures/}}
\input{math_commands}
\newtheorem{proposition}{Proposition}
\makeatletter
\def\thebibliography#1{\par\begin{center}{\bf References}\end{center}\vspace{-0.5\baselineskip}\@mkboth
{REFERENCES}{REFERENCES}\list
{}{\labelwidth 0in\leftmargin\labelwidth
\itemsep .01in
\parsep 0pt
\topsep 0pt
\partopsep 0pt}
\def\newblock{\hskip .11em plus .33em minus .07em}
\sloppy\clubpenalty4000\widowpenalty4000
\sfcode`\.=1000\relax}
\makeatother

\title{PHF: Privileged Hidden Flow for\\ On-Policy Self-Distillation}
\author{
  Yuhan Li\textsuperscript{1},
  Mingxu Zhang\textsuperscript{1},
  Dazhong Shen\textsuperscript{2,*},
  Ying Sun\textsuperscript{3}\thanks{Corresponding authors.} \\
  \textsuperscript{1}The Hong Kong University of Science and Technology (Guangzhou) \\
  \textsuperscript{2}Nanjing University of Aeronautics and Astronautics \\
  \textsuperscript{3}The 63rd Research Institute, National University of Defense Technology, Nanjing\\
  \texttt{yuhanli530@gmail.com},
  \texttt{shendazhong@nuaa.edu.cn}, \\
  \texttt{sunyinggilly@gmail.com}
}

\begin{document}
\maketitle

\begin{abstract}
\input{sections/0_abstract}
\end{abstract}

\begin{figure*}[t]
    \centering
    \makebox[\textwidth][c]{\includegraphics[width=1.06\textwidth]{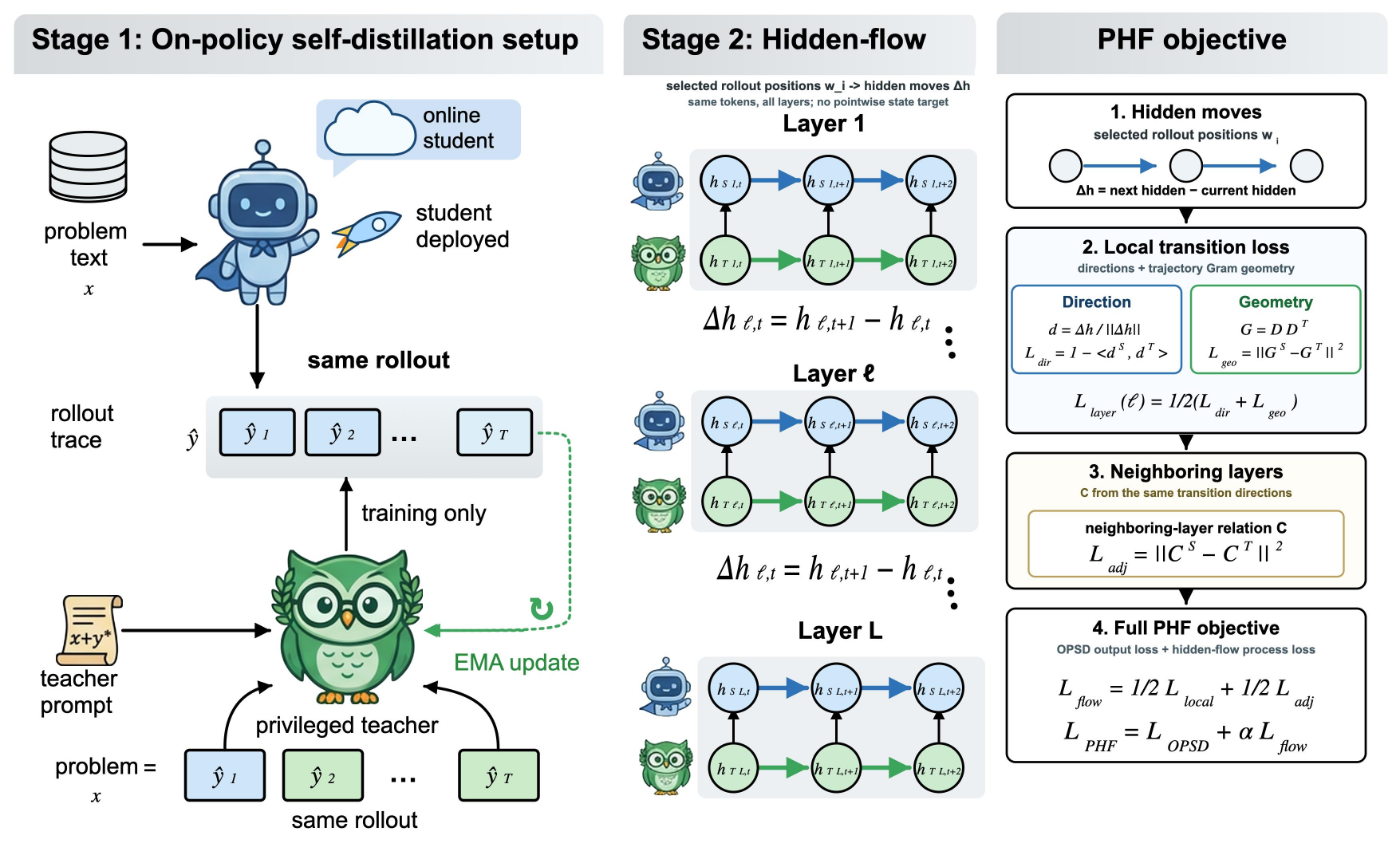}}
    \caption{\textbf{Overview of Privileged Hidden Flow (PHF).}
    The online student samples a rollout from the problem alone. A privileged
    EMA teacher then conditions on the reference solution and the same rollout
    tokens, supplying both the OPSD output-distribution target and the PHF
    hidden-flow target. PHF matches normalized token-to-token hidden transitions
    and their trajectory geometry at every layer, then combines this local
    transition loss with a neighboring-layer relation computed from the same
    transitions. The full objective adds this hidden-flow loss to the standard
    OPSD output loss. The privileged teacher is used only during training, and
    the deployed model is the online student.}
    \label{fig:arch}
\end{figure*}

\input{sections/1_introduction}
\input{sections/2_related_work}
\input{sections/3_method}
\input{sections/4_experiments}
\FloatBarrier
\balance
\input{sections/5_conclusion}

\clearpage
\onecolumn
\begin{multicols}{2}
\bibliographystyle{aaai}
\bibliography{references}
\end{multicols}

\twocolumn
\appendix
\input{sections/A_appendix}

\end{document}

%% file: math_commands.tex
\newcommand{\lossopsd}{\mathcal{L}_{\mathrm{OPSD}}}
\newcommand{\lossflow}{\mathcal{L}_{\mathrm{Flow}}}
\newcommand{\losshtt}{\mathcal{L}_{\mathrm{layer}}}
\newcommand{\lossdir}{\mathcal{L}_{\mathrm{dir}}}
\newcommand{\lossgeo}{\mathcal{L}_{\mathrm{geo}}}
\newcommand{\losslocal}{\mathcal{L}_{\mathrm{local}}}
\newcommand{\lossadj}{\mathcal{L}_{\mathrm{adj}}}
\newcommand{\lossphf}{\mathcal{L}_{\mathrm{PHF}}}
\newcommand{\window}{\mathcal{W}}

%% file: sections/0_abstract.tex
On-policy self-distillation (OPSD) trains a reasoning model on rollouts sampled
from its own policy by matching a privileged teacher that also sees verified
reference solutions. Existing OPSD objectives supervise only the output
distribution, so privileged context affects training through a token-level
divergence without directly supervising the internal computation that produced
that distribution.
We propose Privileged Hidden Flow (PHF), which additionally distills how a
privileged teacher's hidden states move along the same rollout. Rather than
forcing each student hidden vector to match the teacher vector at the same token
position, PHF aligns token-to-token transition directions and trajectory geometry
over selected generated positions.
The all-layer recipe also includes an adjacent-layer relation computed from
these same transitions, without pointwise hidden-state imitation. Under the same
100-step training schedule, PHF improves the Average@12 aggregate over
our reproduced OPSD baseline on Qwen3-1.7B, 4B, and 8B, with observed gains of
about $+2.2$, $+1.5$, and $+1.7$ points. The transport objective is exactly
invariant to shared trajectory offsets; its local geometry term is also
invariant to orthogonal transformations of transition directions.
Ablations distinguish the fixed PHF recipe from pointwise hidden-state matching,
single-channel transition losses, and layer-subset choices, supporting PHF as a
compact hidden-flow extension to OPSD.

%% file: sections/1_introduction.tex
\section{Introduction}

On-policy distillation is a practical recipe for reducing distribution mismatch
in language-model distillation. Instead of imitating a fixed offline corpus, the
student samples its own rollouts and receives dense supervision on the states it
actually visits~\cite{agarwal2023gkd,gu2023minillm,ko2024distillm}.
On-policy self-distillation (OPSD) specializes this idea: a single model acts as
both student and teacher, with the teacher conditioned on privileged information
such as a verified reference solution while the student sees only the
problem~\cite{zhao2026selfdistilled}. This avoids an external teacher and can be
token-efficient for mathematical reasoning.

Existing OPSD objectives, however, capture only one route by which privileged
context changes the model. At a student-generated prefix, conditioning on the
reference can change not only the next-token posterior but also the intermediate
representations that support that posterior. Output-only supervision must express
this privileged information through a token-level divergence at the output head,
without directly constraining the hidden trajectory that produced it.
Process-level supervision has improved reasoning through step-level reward
models~\cite{lightman2023verify}, but at the granularity of output steps rather
than internal dynamics. This raises a direct question:
\emph{can OPSD make better use of privileged context by distilling the internal
process it induces, not only the final output distribution?}

A natural answer is to supervise hidden states directly. But this is harder than
it appears. Hidden coordinates are not semantically fixed: they drift during
training, so an $\ell_2$ target between student and teacher hidden vectors chases
a moving point. Moreover, the hidden computation changes across layers as well
as across generated rollout positions. A per-layer loss alone also ignores
whether adjacent layers preserve similar relations among the same token-to-token
hidden moves.

We introduce \emph{Privileged Hidden Flow} (PHF), which addresses both
difficulties (Figure~\ref{fig:arch}). Rather than matching hidden states
pointwise, PHF matches hidden \emph{transitions}: local displacements of the
residual stream along selected rollout positions. Transition
matching aligns the \emph{direction} and trajectory \emph{geometry} of the
student's internal computation with the privileged teacher's, and does not
require the two to occupy the same point in hidden space. The resulting
objective is exactly invariant to shared offsets of the hidden trajectory, and
its geometry term to orthogonal transformations of the transition
directions (Section~3), which are simple components of representational drift
that pointwise state losses must absorb. We instantiate the privileged teacher
with the same smoothed teacher used by OPSD-style self-distillation, but the
central design choice is the transition target rather than the averaging
mechanism.
In short: OPSD teaches the student \emph{which token} the privileged teacher
would predict; PHF additionally teaches it \emph{how the privileged teacher's
hidden state moves} to get there.
The main recipe uses the same transition object twice: first within each layer
for local direction and geometry matching, and then across neighboring layers to
keep adjacent depth transformations compatible.

PHF adds one scalar loss coefficient to the standard OPSD objective; the
128-position hidden budget and all-layer aggregation recipe are fixed across
scales. It introduces no correctness routing, token filtering, reward model, or
extra rollouts. On the official OPSD math setting evaluated on
American Invitational Mathematics Examination (AIME) 2024, AIME 2025, and
Harvard--MIT Mathematics Tournament (HMMT) 2025, PHF improves the checkpoint-100
Average@12 aggregate over our reproduced OPSD baseline by about $+2.2$ on
Qwen3-1.7B, $+1.5$ on Qwen3-4B, and $+1.7$ on Qwen3-8B. The aggregate gain is
positive at every scale, and all but one benchmark-scale cell moves in the same
direction. Ablations then locate the design boundary: the transition target,
all-layer aggregation, and shared privileged teacher source matter, while nearby
pointwise and selected-layer hidden-supervision controls remain competitive in
some settings.

\noindent\textbf{Contributions.}
\begin{itemize}
\item We introduce a privileged transition-transport objective for OPSD. On
student rollouts, PHF matches normalized token-to-token hidden transitions and
their within-trajectory Gram geometry, rather than matching hidden-state
vectors point by point.
\item We characterize simple invariances of the transport loss. The local
transition objective is unchanged by hidden-trajectory offsets and positive
per-transition rescaling after normalization; its geometry term is also invariant
to independent orthogonal transformations of the student and teacher transition
sets. These structural properties separate PHF from pointwise hidden-state
matching.
\item We average the same transition loss over all layers and, in the main
recipe, add a fixed neighboring-layer relation computed from the same
transitions. This avoids selecting a small hand-picked layer subset while keeping
the supervision tied to hidden motion.
\item We evaluate the fixed recipe on Qwen3-1.7B, 4B, and 8B under the official
OPSD thinking-mode protocol, including Base, SFT, and GRPO context rows. PHF
improves the checkpoint-100 Average@12 aggregate over our reproduced OPSD
baseline at all three scales, and ablations isolate the roles of transition geometry, pointwise
hidden-state controls, and channel coupling.
\end{itemize}

%% file: sections/2_related_work.tex
\section{Related Work}

\paragraph{Distillation and on-policy self-distillation.}
Knowledge distillation transfers a teacher's behavior to a student, from early
compression methods~\cite{bucila2006model,ba2014deep} to softened output
matching~\cite{hinton2015distilling} and sequence-level
distillation~\cite{kim2016sequence}; surveys include
\cite{gou2021survey,xu2024survey}. On-policy variants reduce distribution
mismatch by supervising states visited by the student, as in generalized
knowledge distillation~\cite{ross2011dagger,agarwal2023gkd}, MiniLLM's
reverse-KL training~\cite{gu2023minillm}, and DistiLLM's skewed divergences and
off-policy reuse~\cite{ko2024distillm}. OPSD applies this idea to reasoning
self-distillation: the teacher is the same model conditioned on privileged
information, such as a verified reference, while the student sees only the
problem~\cite{zhao2026selfdistilled}. PHF keeps OPSD's on-policy output
supervision but adds a hidden-process target on the same rollout.

\paragraph{Hidden and relational representation distillation.}
Intermediate-feature distillation includes hidden-layer hints~\cite{romero2015fitnets},
multi-layer transfer for BERT~\cite{sun2019patient}, and attention or embedding
matching in compact language models~\cite{jiao2020tinybert,sanh2019distilbert}.
Related feature losses often match hidden states at selected layers, together
with output, attention, or other representation losses~\cite{yim2017fsp,heo2019overhaul}.
A relational line instead matches structure among examples or features, as in
relational distillation~\cite{park2019relational}, attention
transfer~\cite{zagoruyko2017attention}, and contrastive representation
distillation~\cite{tian2020contrastive}. PHF follows the relational spirit but
changes the object being matched: it compares token-to-token hidden transitions
and their trajectory geometry within a rollout, rather than static hidden
states at individual positions. This sets the closest novelty boundary for PHF
and motivates the pointwise hidden-state and selected-layer controls in our
experiments. We do not claim that hidden representation supervision is new by
itself; the contribution is the specific privileged, on-policy
transition-geometry target and its empirical boundary against nearby pointwise
state and layer-selected controls.

\paragraph{Reasoning and process supervision.}
Reasoning models benefit from chain-of-thought prompting~\cite{wei2022cot},
sampling-based self-consistency~\cite{wang2023selfconsistency}, and training on
rationales or generated solutions~\cite{zelikman2022star,hsieh2023stepbystep,singh2024beyond},
with evaluation on mathematical-reasoning benchmarks~\cite{cobbe2021gsm8k,hendrycks2021math,lewkowycz2022minerva}
and strong math models~\cite{guo2025deepseekr1,yang2024qwen25math}. Process
supervision and reward modeling score intermediate reasoning steps rather than
only final answers~\cite{uesato2022process,lightman2023verify,wang2024mathshepherd},
while policy-optimization methods such as PPO, RLHF, and
GRPO~\cite{schulman2017ppo,ouyang2022instructgpt,shao2024deepseekmath} optimize
action-level behavior through rewards. PHF shares the process-oriented
motivation, but its supervision is internal: it distills how privileged context
changes hidden transitions along the rollout.

%% file: sections/3_method.tex
\section{Method}
\label{sec:method}

\subsection{OPSD Preliminaries}

Let $x$ be a problem, $r$ privileged information such as a reference solution,
and $y_{<t}$ a prefix sampled from the current student. OPSD compares two
conditionals on the same student-visited state:
\begin{align}
p_S(\cdot \mid x, y_{<t}) &= \pi_{\theta}(\cdot \mid x, y_{<t}), \\
p_T(\cdot \mid x, r, y_{<t}) &= \pi_{\bar{\theta}}(\cdot \mid x, r, y_{<t}),
\end{align}
where $\theta$ is the online student and $\bar{\theta}$ the teacher parameters.
Standard OPSD minimizes a per-token divergence, here the Jensen-Shannon
divergence, on student-visited states:
\begin{equation}
\lossopsd
= \frac{1}{T}\sum_{t=1}^{T}
D_{\mathrm{JS}}\!\left(
p_T(\cdot \mid x,r,y_{<t})
\;\|\;
p_S(\cdot \mid x,y_{<t})
\right).
\end{equation}
This objective supervises only the output policy. In practice the per-token
divergence is clipped at a small trust-region value, following the official
OPSD recipe; implementation details are given in the supplementary material.
PHF keeps this output channel unchanged and adds a process-level hidden flow
term.

\subsection{Hidden Transition Transport}

Operationally, one PHF update first samples a student rollout, then runs the
student on the plain prompt and the privileged teacher on the privileged prompt using
the same rollout tokens. The standard OPSD loss matches their output
distributions, while the additional flow loss compares how their hidden states
move along the rollout at each layer. The resulting per-layer losses are then
aggregated across depth.

For layer $\ell$, let $h^S_{\ell,t}$ and $h^T_{\ell,t}$ be the residual-stream
hidden states of the student and privileged teacher at rollout position $t$.
Let $\window=\{w_1<\cdots<w_m\}$ be the deterministic set of valid generated
positions used for the hidden loss. We define the local hidden transition as the
difference between neighboring selected positions along the rollout,
\begin{equation}
\Delta h^S_{\ell,i} = h^S_{\ell,w_{i+1}} - h^S_{\ell,w_i}, \qquad
\Delta h^T_{\ell,i} = h^T_{\ell,w_{i+1}} - h^T_{\ell,w_i},
\end{equation}
and normalize transitions to unit directions,
\begin{equation}
d^S_{\ell,i} =
\frac{\Delta h^S_{\ell,i}}{\|\Delta h^S_{\ell,i}\|_2+\epsilon},
\qquad
d^T_{\ell,i} =
\frac{\Delta h^T_{\ell,i}}{\|\Delta h^T_{\ell,i}\|_2+\epsilon}.
\end{equation}
The process axis is the generated rollout position; transformer depth is an
observation axis over which we repeat the same comparison. PHF does not use a
layer-to-layer difference such as $h_{\ell+1,t}-h_{\ell,t}$ as its basic
transition object.
The transition-direction loss aligns local displacements over this selected
position set,
\begin{equation}
\lossdir(\ell)
=
\frac{1}{m-1}
\sum_{i=1}^{m-1}
\bigl(1-\langle d^S_{\ell,i}, d^T_{\ell,i}\rangle\bigr).
\end{equation}
Direction matching aligns each step but ignores the \emph{shape} of the
trajectory. We therefore add a transition-geometry term. Let $D^S_\ell$ stack
the rows $d^S_{\ell,i}$ for $i=1,\ldots,m-1$, and likewise $D^T_\ell$; define the
Gram matrices $G^S_\ell = D^S_\ell (D^S_\ell)^\top$ and
$G^T_\ell = D^T_\ell (D^T_\ell)^\top$. The geometry loss matches which
transitions are parallel, orthogonal, or opposing:
\begin{equation}
\lossgeo(\ell)
=
\frac{1}{(m-1)^2}
\bigl\|G^S_\ell-G^T_\ell\bigr\|_F^2 .
\end{equation}
The Gram comparison is a within-sequence, representation-similarity-style
construction inspired by CKA~\cite{kornblith2019similarity}, applied to
transition directions rather than states: it supervises the relational
structure of the trajectory rather than pointwise hidden coordinates.
The per-layer transition loss is the mean of the two,
\begin{equation}
\losshtt(\ell)
=
\tfrac{1}{2}\bigl(\lossdir(\ell)+\lossgeo(\ell)\bigr).
\end{equation}
This differs from pointwise hidden-state matching in \emph{what is asked of the
student}: not to make $h^S_{\ell,t}$ equal $h^T_{\ell,t}$ at every selected
position, but to make comparable token-to-token moves under the privileged
process. The next subsection makes the resulting invariances precise.

\paragraph{Neighboring-layer relation.}
As part of all-layer aggregation, PHF also compares how adjacent layers relate
the same transition directions.
A purely per-layer average does not check whether adjacent layers relate these
motions in the same way as the privileged teacher. PHF therefore uses the same
transition-only object to compare neighboring layers. Define
\begin{equation}
C^S_{\ell,\ell+1}
=
D^S_\ell (D^S_{\ell+1})^\top,
\qquad
C^T_{\ell,\ell+1}
=
D^T_\ell (D^T_{\ell+1})^\top .
\end{equation}
These matrices measure how each selected rollout move at layer $\ell$ relates to
the moves one layer later. PHF matches this relation between the student and the
privileged teacher,
\begin{equation}
\lossadj
=
\frac{1}{L-1}\sum_{\ell=1}^{L-1}
\frac{1}{(m-1)^2}
\bigl\|C^S_{\ell,\ell+1}-C^T_{\ell,\ell+1}\bigr\|_F^2 .
\end{equation}
This term is still transition-based: it never asks the student to match a
teacher hidden vector at a fixed token position. Instead, it asks the student to
pass hidden motions between neighboring layers with the same relational pattern
as the privileged teacher.

\subsection{Structural Properties}
\label{sec:invariance}

The local transport objective has exact algebraic invariances that pointwise
hidden-state matching lacks. Fix a layer $\ell$ and write
$d^S_{\ell,i},d^T_{\ell,i}$ for the normalized student and teacher transition
directions, with Gram matrices $G^S_\ell$ and $G^T_\ell$.

\begin{proposition}[Invariances of transition transport]
\label{prop:invariance}
For any layer $\ell$, $\lossdir(\ell)$ and $\lossgeo(\ell)$ are unchanged by:
\begin{enumerate}
    \item adding any position-independent offset to either trajectory,
    $h^S_{\ell,t}\mapsto h^S_{\ell,t}+c^S_\ell$ and
    $h^T_{\ell,t}\mapsto h^T_{\ell,t}+c^T_\ell$;
    \item multiplying any transition by a positive scalar before normalization,
    $\Delta h^{S/T}_{\ell,i}\mapsto s^{S/T}_{\ell,i}\Delta h^{S/T}_{\ell,i}$,
    up to the numerical $\epsilon$ in the denominator.
\end{enumerate}
In addition, for any orthogonal maps $R_S,R_T$, the geometry loss is unchanged
under $d^S_{\ell,i}\mapsto R_S d^S_{\ell,i}$ and
$d^T_{\ell,i}\mapsto R_T d^T_{\ell,i}$; the direction loss has the same
orthogonal invariance when $R_S=R_T$. In contrast, the pointwise state loss
$\|h^S_{\ell,t}-h^T_{\ell,t}\|_2^2$ generally changes under these transformations.
\end{proposition}

The proof follows from differencing, normalization, and the identity
$(DR)(DR)^\top=DD^\top$ for orthogonal $R$. These properties give the intended
reading of PHF: the local loss tracks \emph{how} the representation moves while
discarding its absolute location and normalized step magnitude. The geometry
term further compares the relational shape of the transition trajectory rather
than pointwise hidden coordinates. The neighboring-layer term uses the same
transitions and retains relative orientation between adjacent layers because
that relative orientation is the compatibility signal it measures.

\paragraph{Connection to the output channel.}
The two supervision channels are algebraically linked at the final layer.
Writing $\tilde h_{L,t}$ for the final-norm output and $W$ for the linear LM
head, the logit displacement between adjacent positions is
$\Delta z_t = W(\tilde h_{L,t+1}-\tilde h_{L,t})$: the output posterior can
change only through the head's image of the final-layer transition.
Output-level OPSD therefore already supervises a \emph{projection} of one
transition. The flow term adds targets on how the privileged state moves from
one position to the next at each layer, including hidden directions that are not
directly specified by the scalar output-posterior divergence. In this sense PHF
adds a process signal connected to OPSD's output channel rather than an unrelated
auxiliary objective. We do not claim that this quantifies independent
information; the coupling ablations in Section~4 only probe whether sharing the
same privileged process source matters empirically. This link should not be read
as an information-theoretic certificate. We do not show that hidden flow carries
information beyond what is recoverable from the output posterior; the hidden-flow
loss is an auxiliary training target whose value is tested under the fixed OPSD
protocol.

\subsection{Privileged Teacher Source}

We instantiate the privileged teacher as an EMA copy of the online model,
following the standard use of weight-averaged teachers to smooth training
targets~\cite{polyak1992acceleration,tarvainen2017mean},
\begin{equation}
\bar{\theta}_{k}
=
\rho\,\bar{\theta}_{k-1}
+
(1-\rho)\,\theta_k,
\qquad \rho=0.999 ,
\end{equation}
used only during training. Both its output logits and hidden transitions are
computed under the privileged prompt $(x,r,y_{<t})$ and detached from the
gradient. This averaging choice stabilizes the target under a changing
on-policy state distribution; it is not a deployment mechanism, and the
evaluated model is always the online student $\theta$.

\subsection{All-Layer Aggregation}

Hidden distillation usually depends on selecting a few intermediate layers. PHF
instead uses all layers. The local PHF component averages
the per-layer transition loss over depth,
\begin{equation}
\losslocal
=
\frac{1}{L}\sum_{\ell=1}^{L}
\losshtt(\ell).
\end{equation}
The PHF recipe used for the main table gives equal weight to motion matching
along selected rollout positions inside each layer and a neighboring-layer
relation computed from the same selected-position transitions,
\begin{equation}
\lossflow
=
\tfrac{1}{2}\losslocal+\tfrac{1}{2}\lossadj .
\end{equation}
This removes the layer-selection knob; the neighboring-layer relation is
a fixed aggregation choice that avoids treating layers as fully independent
curves. The method has one coefficient $\alpha$ for the whole process channel;
the half-and-half split is fixed across scales and should be read as part of the
evaluated recipe rather than as an optimized weighting rule.

\subsection{Full Objective}

The final PHF objective adds the flow term to OPSD,
\begin{equation}
\lossphf
=
\lossopsd
+
\alpha\,\lossflow ,
\end{equation}
where $\lossflow=\losslocal$ for the local transition version and
$\lossflow=\frac12\losslocal+\frac12\lossadj$ for PHF with neighboring-layer
relations, and
$\alpha=0.05$, chosen once to be on the same order as the clipped
token-level OPSD loss. We keep this coefficient fixed across model scales rather
than tuning it per model. The flow is computed over at most
$|\window|=128$ valid generated positions, selected deterministically to cover
the rollout. This is a compute-driven truncation because storing hidden states
for every layer over full rollouts is memory-prohibitive. We did not sweep the
position budget or sampling rule; the 128-position choice is part of the recipe
evaluated here, not a claim that other positions are uninformative. This is the
only tuned scalar loss weight PHF adds; the EMA teacher, position budget, and
local/neighboring-layer split are fixed recipe choices. No correctness routing,
token filtering, reward model, or additional
rollout set is introduced.

%% file: sections/4_experiments.tex
\section{Experiments}

\subsection{Setup}

\paragraph{Models and data.}
We evaluate on the official OPSD mathematical-reasoning setting with three
Qwen3 models~\cite{qwen3technical}: Qwen3-1.7B, Qwen3-4B, and Qwen3-8B. Their
layer counts and hidden sizes are taken from the released model configurations:
28 layers with hidden size 2048 for 1.7B, and 36 layers with hidden sizes 2560
and 4096 for 4B and 8B. Training
uses the official OPSD math set of $29{,}434$ examples with student rollouts of
length $1024$.

\paragraph{Training.}
For the OPSD/PHF comparison, we follow the official OPSD
learning-rate schedule (cosine decay) and run on-policy training for $100$
optimizer steps with LoRA and rollout temperature $1.1$. The JSD token clip is
$0.05$ for Qwen3-1.7B and Qwen3-4B, and $0.06$ for Qwen3-8B, matching the
reproduced OPSD configuration at each scale. PHF adds hidden transition
supervision with window
$|\window|{=}128$, coefficient $\alpha{=}0.05$, and EMA decay $\rho{=}0.999$.
OPSD and PHF use the same schedule and step budget. PHF keeps the OPSD output
loss but adds the privileged hidden-process recipe described in
the Method section. The privileged forward pass conditions on verified
reference solutions during training only; evaluation always uses the online
student without references. Teacher-source controls are reported separately.
Base, SFT, and GRPO rows are context
baselines evaluated with the same thinking-mode Average@12 protocol.

\paragraph{Evaluation.}
We use the official OPSD thinking-mode evaluation on the American Invitational
Mathematics Examination 2024, American Invitational Mathematics Examination
2025, and Harvard--MIT Mathematics Tournament 2025 (30 problems each). For every
problem we draw $n{=}12$ samples
(temperature $1.0$, top-$p{=}0.95$, up to $38{,}912$ new tokens, thinking
enabled) and report Average@12, the mean per-sample correctness. The primary
aggregate is the reported Average column for the evaluation suite. Unless stated
otherwise, every reported trained checkpoint is evaluated at checkpoint~100.

\paragraph{Baselines.}
Our principal reference point is the reproduced \emph{OPSD} baseline
(output-only, same schedule and budget), since PHF is an additive extension of
the same framework. We additionally include the base
checkpoint, a supervised-finetuning checkpoint, and a GRPO checkpoint as context for
the scale of the OPSD/PHF gains.

\subsection{Main Results}

\input{tables/table1_main}

Table~\ref{tab:main} reports the reproduced checkpoint-100 comparison. PHF
improves the Average@12 aggregate over the OPSD baseline at all three model
scales: about $+2.2$ points (1.7B), $+1.5$ points (4B), and $+1.7$ points
(8B). With the same hidden window and PHF coefficient, the primary aggregate moves in
the same direction at every size; all but one reported benchmark-scale cell also
has a positive observed checkpoint-100 delta. These deltas show that an
auxiliary process-level signal can complement an already strong
output-distribution objective, not replace it.

\paragraph{Interpreting the comparison.}
The cross-scale comparison is the main evidence surface: within the reproduced
OPSD setting, the fixed PHF recipe combines a transition-based process target,
trajectory geometry, and all-layer aggregation into a consistently positive
OPSD extension across the three tested model scales.
The context rows are intentionally not treated as alternative PHF baselines:
Base, SFT, and GRPO use the same evaluation harness, but they differ in training
objective or supervision source. The controlled comparison is therefore OPSD
versus PHF at matched step budget, rollout distribution, checkpoint, LoRA
configuration, and evaluation protocol. This keeps the evidence focused on the
incremental hidden-process target rather than on differences in data, reward
modeling, or decoding.

\subsection{Design Ablations}
\label{sec:ablations}

Table~\ref{tab:ablations} reports the design ablations that were evaluated at
all three model scales. Each row keeps the OPSD output objective, training
schedule, checkpoint, and 128-position hidden budget fixed unless the row name
states the changed factor.

\input{tables/table3_ablations}

Three patterns emerge. First, PHF is the strongest aggregate row at every scale.
Second, direction-only and geometry-only variants expose useful signal, but
neither component alone matches the full direction-plus-geometry recipe across
scales. Third, the selected-layer control trails all-layer PHF, supporting the
choice to average the transition objective across depth rather than choosing a
small hand-picked layer subset.
The ablation table is deliberately narrow: it includes only variants evaluated
for all three model sizes under the same checkpoint-100 protocol. Rows with
partial scale coverage are kept out of the main table because they are useful
diagnostics but would weaken the comparison surface. Within this restricted
surface, the main recipe is not just a larger loss; it preserves the same OPSD
output objective and changes only how the privileged hidden process is matched.

\subsection{Analysis}

\paragraph{Why transitions rather than states.}
Pointwise hidden-state matching (as in normalized MSE objectives) requires the
student and teacher to occupy the same point in representation space at each
selected token position, a target that itself moves as the student's LoRA
updates shift the manifold during on-policy training. Transition transport asks
only that the student \emph{move} like the teacher: by
Proposition~\ref{prop:invariance} it is invariant to trajectory offsets and
insensitive to transition rescaling after normalization, and its geometry term
to independent rotations of the two direction sets. These are common components
of representational drift that pointwise MSE must absorb as loss. The Gram term
additionally targets
relational trajectory structure (which transitions are parallel, orthogonal, or
opposing) rather than a pointwise state target.
Table~\ref{tab:ablations} tests the transition components across all three model
scales. Keeping only the direction term loses the aggregate gain at 1.7B and
8B, while keeping only the geometry term is more competitive but still trails
the full PHF recipe at every scale. These rows indicate that the two components
are not interchangeable: direction matching supplies aligned local motion,
whereas the Gram term supplies relational trajectory shape. Thus single-channel
controls expose useful signal, but they do not give the same transition-based
process objective as the fixed direction-plus-geometry recipe.
This distinction matters because the privileged reference changes the
\emph{process} the model is asked to imitate, not merely the final answer
distribution. Direction matching aligns local motion between adjacent selected
positions, while the geometry term constrains the relations among those motions
inside the rollout. PHF therefore supplies a target that is invariant to simple
coordinate shifts but still sensitive to whether the student's hidden trajectory
organizes the same reasoning steps as the privileged teacher.

\paragraph{Role of the EMA teacher.}
On-policy training changes both the model and the sampled state distribution, so
a privileged forward pass from the instantaneous model is a noisy process
estimate. The EMA teacher smooths this process over recent history while still
tracking the student. This does not turn EMA into an independent deployment
mechanism: the online student is always the evaluated model, and the EMA copy is
used only as a training-time process source.
Using the EMA copy as the hidden teacher also keeps the method aligned with the
OPSD deployment story: the evaluated model is the online student, and no
teacher, verifier, or privileged reference is needed at test time. The teacher
is only a stabilized view of the current learner under the reference-conditioned
prompt, so the training signal stays on-policy rather than becoming a separate
offline distillation stage.

\paragraph{Training budget and stability.}
PHF trains under the same optimizer schedule as OPSD without requiring a reward
model, correctness routing, or additional rollout set. The control variants in
Table~\ref{tab:ablations} therefore do not introduce extra rollouts or rewards,
but these results alone do not separate optimization effects from the objective
being tested.
PHF adds one privileged hidden-process forward pass and hidden-state storage,
while keeping the rollout set and optimizer schedule unchanged.
This accounting is important for interpreting the gains. The method does not
increase the number of sampled completions, does not filter trajectories by
correctness, and does not add a reward model. Its extra cost is concentrated in
the hidden-state window and the privileged forward pass used to form the process
target. The comparison therefore asks whether the same on-policy training run can
use the verified reference more effectively, rather than whether additional
search or reward feedback improves the final answer.

\subsection{Checkpoint Dynamics}
\label{sec:dynamics}

\begin{figure*}[!t]
    \centering
    \makebox[\textwidth][c]{\includegraphics[width=1.04\textwidth]{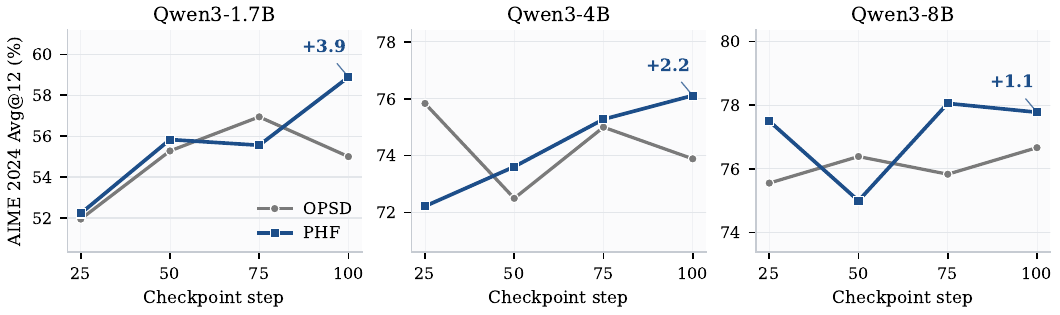}}
    \caption{AIME 2024 Average@12 checkpoint dynamics for OPSD and PHF, with one
    panel per model scale. The checkpoint-100 markers match the main
    Table~\ref{tab:main} results.}
    \label{fig:eval-dynamics}
\end{figure*}

Figure~\ref{fig:eval-dynamics} plots available AIME 2024 Average@12 checkpoint
evaluations. The curves show checkpoint trajectories for the corresponding
OPSD/PHF recipe, and the checkpoint-100 markers are aligned to the main
Table~\ref{tab:main} results. The plot is not an additional selection rule: all
reported headline numbers still come from the fixed checkpoint-100 protocol.
The curves also illustrate why we report the checkpoint-100 table rather than a
best-checkpoint sweep. Intermediate checkpoints can move non-monotonically under
the stochastic thinking-mode evaluation, especially with $n{=}12$ samples per
problem. We therefore use the dynamics plot only as a sanity check that PHF is
being compared along the same training trajectory family as OPSD, not as a
criterion for choosing a more favorable endpoint.

\paragraph{All-layer spread.}
The recorded all-layer diagnostics indicate that the hidden-flow layer energy is
not concentrated in a single layer in these diagnostic runs. At checkpoint~100,
the largest layer contributes $6.5\%$, $5.1\%$, and $4.8\%$ of the recorded
layer energy for Qwen3-1.7B, Qwen3-4B, and Qwen3-8B, respectively. The
corresponding layer-energy entropies are $3.17$, $3.38$, and $3.40$ (with
maxima $\log 28$ and $\log 36$). We use these trainer-state diagnostics only to
justify all-layer aggregation as a reasonable default, not as evidence that the
hidden loss carries independent behavioral information beyond the output
posterior.
The selected-layer control in Table~\ref{tab:ablations} is consistent with this
view: a fixed hand-picked subset is a useful boundary check, but it leaves
performance below the all-layer PHF recipe at every tested scale. We therefore
treat all-layer aggregation as part of the fixed method, not as a post-hoc
selection over layers.

\subsection{Discussion}

\paragraph{What the hidden-flow target adds.}
PHF is best read as a way to use the same privileged reference more
structurally. OPSD already distills the privileged output distribution at
student-visited states; PHF asks whether the student's hidden trajectory moves
through those states in the same local directions and with the same relational
geometry as the privileged teacher. This is a weaker and more stable requirement
than matching hidden vectors point by point, because it does not require the two
models to share an absolute coordinate origin at every selected position. It is
also more informative than a scalar reward, because it supplies a process target
for the path between generated tokens rather than only for the final answer.

\paragraph{Why the comparison is conservative.}
All reported PHF gains are obtained without changing the rollout set, evaluation
protocol, optimizer schedule, checkpoint, or output objective. The method also
does not introduce correctness routing or extra answer sampling. This matters
because many improvements in mathematical reasoning can be explained by more
search, stronger filtering, or additional reward feedback. Here the evidence is
narrower but cleaner: under the reproduced OPSD harness, adding a fixed
transition-based hidden-process loss improves the Average@12 aggregate at every
tested scale.

\paragraph{What the ablations rule out.}
The controls do not show that every hidden supervision signal helps. They show
that the particular PHF recipe is more reliable than the tested single-channel
or selected-layer alternatives under the same three-scale protocol. Direction
only and geometry only each preserve part of the process signal, but neither
matches the full recipe across scales. Selected layers avoids all-layer storage
for some layers but gives up a consistent part of the gain, suggesting that the
privileged process signal is distributed rather than isolated to one manually
chosen depth.

%% file: tables/table1_main.tex
\begin{table*}[t]
\centering
\setlength{\tabcolsep}{0pt}
\renewcommand{\arraystretch}{1.0}
\begin{tabular*}{0.72\textwidth}{@{\extracolsep{\fill}}llrrrr@{}}
\toprule
Scale & Method & AIME24 & AIME25 & HMMT25 & Average \\
\midrule
\multirow{5}{*}{1.7B}
 & Base & 48.06 & 36.94 & 22.50 & 35.83 \\
 & SFT & 45.00 & 36.11 & 21.39 & 34.17 \\
 & GRPO & 48.06 & 35.28 & 21.67 & 35.00 \\
 & OPSD & 55.00 & 40.83 & 26.67 & 40.83 \\
 & \textbf{PHF} & \textbf{58.89} & \textbf{41.67} & \textbf{28.61} & \textbf{43.06} \\
\midrule
\multirow{5}{*}{4B}
 & Base & 74.17 & 66.67 & 40.06 & 60.30 \\
 & SFT & 70.00 & 66.11 & 35.56 & 57.22 \\
 & GRPO & 74.17 & 66.94 & 43.21 & 61.44 \\
 & OPSD & 73.89 & \textbf{68.33} & 43.45 & 61.89 \\
 & \textbf{PHF} & \textbf{76.11} & 68.06 & \textbf{46.00} & \textbf{63.39} \\
\midrule
\multirow{5}{*}{8B}
 & Base & 74.44 & 66.67 & 43.06 & 61.39 \\
 & SFT & 73.89 & 65.83 & 37.22 & 58.98 \\
 & GRPO & 76.39 & 68.33 & 44.72 & 63.15 \\
 & OPSD & 76.67 & 68.33 & 47.78 & 64.26 \\
 & \textbf{PHF} & \textbf{77.78} & \textbf{71.72} & \textbf{48.33} & \textbf{65.94} \\
\bottomrule
\end{tabular*}
\caption{Checkpoint-100 Average@12 on Qwen3 models under thinking-mode
evaluation. Average is the suite aggregate.
Base, SFT, and GRPO are context rows under the same $n{=}12$ protocol; the
primary comparison is OPSD vs.\ PHF under the same 100-step schedule.}
\label{tab:main}
\end{table*}

%% file: tables/table3_ablations.tex
\begin{table}[t]
\centering
\scriptsize
\setlength{\tabcolsep}{4pt}
\renewcommand{\arraystretch}{0.92}
\resizebox{\columnwidth}{!}{%
\begin{tabular}{lrrr}
\toprule
Variant & 1.7B & 4B & 8B \\
\midrule
OPSD baseline & 40.83 & 61.89 & 64.26 \\
\textbf{PHF} & \textbf{43.06} & \textbf{63.39} & \textbf{65.94} \\
\midrule
Direction only & 39.72 & 63.06 & 63.70 \\
Geometry only & 41.67 & 62.59 & 65.74 \\
Selected layers & 41.39 & 62.87 & 65.19 \\
\bottomrule
\end{tabular}}
\caption{Cross-scale checkpoint-100 design ablations. Entries are Average@12
suite aggregates. All rows are evaluated at all three model scales under the
same OPSD output objective, training schedule, checkpoint, and 128-position
hidden budget.}
\label{tab:ablations}
\end{table}

%% file: sections/5_conclusion.tex
\section{Conclusion}

We introduced \emph{Privileged Hidden Flow} (PHF), a process-level extension of
on-policy self-distillation that distills both the privileged output
distribution and the hidden transition dynamics induced by privileged context.
PHF matches transition directions and trajectory geometry rather than pointwise
hidden-state vectors, yielding offset invariance, transition-scale insensitivity
after normalization, and (for its geometry term) orthogonal invariance. Its
targets come from a privileged teacher used strictly as a training-time source;
the fixed recipe also compares how neighboring layers transform the same
selected-position changes. Across Qwen3-1.7B, 4B, and 8B checkpoint-100
evaluations, PHF improves the Average@12 aggregate over the OPSD
baseline at every scale on competition mathematics, with gains of about $+2.2$,
$+1.5$, and $+1.7$ points. Ablations delimit the design: coupling the hidden
signal to the same privileged output process is supported by the main controls,
while pointwise hidden-state and selected-layer variants remain useful boundary
checks.
The main takeaway is that privileged information can supervise more than the
next-token distribution. When the reference solution is available during
training, it also induces a hidden process for how the model moves through the
student's own rollout. PHF turns that process into a local transition target
while preserving the deployment setting of OPSD: the online student is evaluated
alone, without references, teachers, verifiers, or extra test-time search. The
evidence is intentionally scoped to a fixed recipe and a matched harness, but it
suggests that process-level distillation can be made compatible with on-policy
self-distillation rather than treated as a separate supervision regime.
This framing also clarifies why the reported effect is modest but meaningful:
PHF is not a new training pipeline, a new verifier, or a larger inference
budget. It is an additional way to spend the information already present in the
verified reference during the same on-policy update. The consistent positive
aggregate shift across 1.7B, 4B, and 8B therefore supports the central design
claim even though the method remains deliberately close to OPSD.

\paragraph{Limitations.}
PHF currently assumes verified reference solutions during training and therefore
inherits the reference quality requirements of OPSD. The method also stores
hidden states for the selected 128-position window, so longer windows require
more memory. The internal controls suggest that hidden-process targets can be
allocated differently across model depth and rollout positions, motivating
broader studies of adaptive depth and token budgets. Our evidence is also
limited to one model family, one mathematical-reasoning training set, and
checkpoint-100 evaluations under the official thinking-mode protocol. We
therefore treat the reported gains as matched-harness point estimates rather
than field-wide claims. The ablations identify the useful parts of the fixed PHF
recipe, but they do not exhaust the design space of hidden-state selection,
teacher smoothing, token windows, or layer weighting.
Average@12 also measures per-sample correctness under a fixed sampling budget,
not proof quality, calibration, or robustness to prompt changes. The paper
therefore does not claim that hidden-flow matching improves every aspect of
reasoning behavior; it shows that, under this protocol, transition-level
privileged supervision improves the reported competition-math aggregate.

\paragraph{Future work.}
Future work should test other model families and non-math tasks and study
adaptive token or layer budgets that reduce memory while preserving the
hidden-process signal. Another useful direction is to learn where the hidden
target should be applied during a rollout, instead of fixing a uniform
128-position window. Finally, PHF should be tested with longer training horizons
and larger-scale reference corpora to separate short-run process alignment from
long-run policy improvement.
Better diagnostics could also compare hidden-flow alignment with solution-level
properties such as proof structure, error type, and recovery from wrong partial
plans, which would clarify when the hidden process target contributes beyond the
output posterior alone.

%% file: sections/A_appendix.tex
\section{Implementation Details}
\label{app:impl}

\paragraph{Models.} We use Qwen3-1.7B (28 layers, hidden size $2048$), Qwen3-4B
(36 layers, $2560$), and Qwen3-8B (36 layers, $4096$). All training uses LoRA
adapters with rank $r{=}64$, $\alpha{=}128$, applied to the attention projections
$\{q,k,v,o\}$ and the MLP projections $\{\mathrm{gate},\mathrm{up},\mathrm{down}\}$.

\paragraph{Optimization.} We follow the official OPSD schedule: AdamW with
learning rate $5\times10^{-6}$, cosine decay scheduled over
\texttt{num\_train\_epochs}$=30$, gradient clipping at $0.1$, per-device batch
size $4$ with gradient accumulation $1$ across $8$ GPUs (effective batch $32$).
A watcher stops training at optimizer step $100$; OPSD and PHF use the same
schedule and step budget. PHF keeps the OPSD output objective and adds the
privileged hidden transition recipe described in the Method section.

\paragraph{Rollouts.} The student samples on-policy rollouts of length $1024$ with
temperature $1.1$, top-$p\,{=}\,0.95$, and top-$k\,{=}\,20$. The output OPSD loss
is a per-token Jensen-Shannon divergence with token-level clipping at $0.05$
for Qwen3-1.7B and Qwen3-4B, and $0.06$ for Qwen3-8B,
between the student conditional $p_S(\cdot\mid x,y_{<t})$ and the privileged
teacher conditional $p_T(\cdot\mid x,r,y_{<t})$, evaluated on student-visited
states.

\paragraph{Hidden flow.} For the local flow term we select up to
$|\mathcal{W}|{=}128$ valid generated positions deterministically across the
rollout, form differences between neighboring selected positions at every
layer, normalize them to unit directions, and average the direction loss
(cosine) with a geometry loss that compares the pairwise relations among those
transitions (squared Frobenius distance between transition Gram matrices). PHF
also has a neighboring-layer term that compares the relation between transitions at
layer $\ell$ and layer $\ell{+}1$ on the same selected position set. The local version uses
$\lossflow=\losslocal$, while the main PHF variant uses
$\lossflow=\frac12\losslocal+\frac12\lossadj$. The privileged teacher is an EMA
copy of the online model with decay $\rho{=}0.999$, computed under the
privileged prompt $(x,r,y_{<t})$ and detached from gradients. The flow
coefficient is $\alpha{=}0.05$ for all scales. No correctness
routing, token filtering, reward model, or extra rollout set is used.

\paragraph{Compute.} All experiments run on a single node of $8{\times}$NVIDIA
H20 GPUs ($143$\,GB each).

\section{Evaluation Protocol}
\label{app:eval}

We evaluate with the official thinking-mode protocol on AIME 2024, AIME 2025, and
HMMT 2025 ($30$ problems each). For each problem we draw $n{=}12$ samples with
temperature $1.0$, top-$p\,{=}\,0.95$, top-$k$ disabled, and up to $38{,}912$ new
tokens, with thinking mode enabled. \textbf{Average@12} is the mean per-sample
correctness over the $12$ samples (equivalently, $\mathrm{pass}@1$ averaged over
samples). The primary aggregate is the reported Average column for the
evaluation suite. The online student $\theta$ is always the evaluated model; the EMA
teacher is never deployed.

\paragraph{Context rows in Table~\ref{tab:main}.}
Base, SFT, and GRPO are protocol-matched context rows, not the main
step-budget-matched comparison. The Base row evaluates the released Qwen3
checkpoint directly. The SFT row is a LoRA supervised-finetuning checkpoint
trained on the same OPSD math data format with reference solutions as target
responses, a maximum sequence length of $16{,}000$, learning rate $5\times10^{-6}$,
effective batch size $32$, and checkpoint~100 evaluation. The GRPO row is a
LoRA policy-optimization checkpoint trained with the same math-answer reward
used by the reproduced OPSD codebase, learning rate $5\times10^{-6}$, effective
batch size $32$, $8$ sampled completions per prompt during training, reward
normalization within group, $\beta{=}0$, and checkpoint~100 evaluation. All
three context rows are evaluated with the same thinking-mode Average@12 protocol
as OPSD and PHF. Only the OPSD and PHF rows should be read as the controlled
same-schedule comparison, because PHF is an additive modification to OPSD.

\section{Hyperparameters}
\label{app:hparams}

\refstepcounter{table}\label{tab:hparams}
\begin{center}
{\small
\setlength{\tabcolsep}{3pt}
\begin{tabular}{@{}p{0.34\columnwidth}p{0.58\columnwidth}@{}}
\toprule
Component & Setting \\
\midrule
LoRA rank / $\alpha$ & $64$ / $128$ \\
LoRA targets & $q,k,v,o,\mathrm{gate},\mathrm{up},\mathrm{down}$ \\
Learning rate & $5\times10^{-6}$ (cosine) \\
Grad clip & $0.1$ \\
Effective batch & $32$ \\
Optimizer steps & $100$ \\
Rollout length & $1024$ \\
Rollout temp. / top-$p$ / top-$k$ & $1.1$ / $0.95$ / $20$ \\
JSD token clip & $0.05$ (1.7B/4B), $0.06$ (8B) \\
Flow coefficient $\alpha$ & $0.05$ \\
EMA decay $\rho$ & $0.999$ \\
Hidden window $|\mathcal{W}|$ & $128$ \\
Layer aggregation & all layers; neighboring layer pairs for main PHF variant \\
\midrule
Eval samples $n$ & $12$ \\
Eval temp. / top-$p$ / top-$k$ & $1.0$ / $0.95$ / disabled \\
Eval max new tokens & $38{,}912$ \\
\bottomrule
\end{tabular}
\par\smallskip
\parbox{0.94\columnwidth}{\centering Table~\thetable: Full PHF training and evaluation hyperparameters.\par}
}
\end{center}

\section{Notation}
\label{app:notation}

\refstepcounter{table}\label{tab:notation}
\begin{center}
{\small
\setlength{\tabcolsep}{3pt}
\begin{tabular}{@{}p{0.32\columnwidth}p{0.60\columnwidth}@{}}
\toprule
Symbol & Meaning \\
\midrule
$x,\ r,\ y$ & problem, privileged reference, student rollout \\
$\theta,\ \bar\theta$ & online student, EMA privileged teacher \\
$p_S,\ p_T$ & student / privileged-teacher output distributions \\
$h^{S}_{\ell,t},\ h^{T}_{\ell,t}$ & hidden state at layer $\ell$, position $t$ \\
$\Delta h_{\ell,i}$ & transition between selected positions \\
$d_{\ell,i}$ & unit-normalized selected-position transition \\
$G_\ell$ & Gram matrix of transition directions \\
$\lossopsd$ & output (JSD) loss \\
$\lossdir,\ \lossgeo$ & transition direction / geometry loss \\
$\losshtt(\ell)$ & per-layer transition loss \\
$\losslocal,\ \lossadj$ & local transition loss / neighboring layer relation loss \\
$\lossflow$ & hidden flow loss ($\losslocal$ or $\frac12\losslocal+\frac12\lossadj$) \\
$\lossphf$ & full objective $\lossopsd + \alpha\lossflow$ \\
$\alpha,\ \rho$ & flow coefficient ($0.05$), EMA decay ($0.999$) \\
$L,\ |\window|$ & number of layers, hidden window ($128$) \\
$n$ & evaluation samples per problem ($12$) \\
\bottomrule
\end{tabular}
\par\smallskip
\parbox{0.94\columnwidth}{\centering Table~\thetable: Notation used throughout the paper.\par}
}
\end{center}

\section{Proof of Proposition~\ref{prop:invariance}}
\label{app:proofs}

Fix a layer and write
$\Delta h_i=h_{w_{i+1}}-h_{w_i}$,
$d_i=\Delta h_i/(\|\Delta h_i\|_2+\epsilon)$, and
$G=DD^\top$, where $D$ stacks the selected transition directions. The local
terms are $\lossdir=\frac{1}{m-1}\sum_i(1-\langle d^S_i,d^T_i\rangle)$ and
$\lossgeo=\frac{1}{(m-1)^2}\|G^S-G^T\|_F^2$. The exact statements below take
$\epsilon=0$; nonzero $\epsilon$ only perturbs the scale statement through the
normalizer.

\paragraph{(i) Offset invariance.}
Adding any position-independent offset $c$ gives
$(h_{w_{i+1}}+c)-(h_{w_i}+c)=\Delta h_i$, so directions, Gram matrices, and both
losses are unchanged. Pointwise matching instead changes with
$\|c_S-c_T\|$ whenever the student and teacher offsets differ.

\paragraph{(ii) Per-step scale invariance.}
For $s_i>0$, $s_i\Delta h_i/\|s_i\Delta h_i\|_2=d_i$; thus unit normalization
removes per-transition magnitude before either local loss is computed. A
pointwise state loss has no analogous invariance because it acts on the states
themselves.

\paragraph{(iii) Orthogonal invariance of the geometry term.}
For orthogonal $R_S$, replacing $D^S$ by $D^S R_S^\top$ leaves
$(D^S R_S^\top)(D^S R_S^\top)^\top=G^S$, and the same holds for an independent
$R_T$ on the teacher side; hence $\lossgeo$ is invariant to independent
orthogonal maps. The direction term is invariant only to a common map
$R_S=R_T$, which preserves each inner product. Thus PHF separates a
coordinate-anchored direction signal from a relational geometry signal, while
pointwise state matching generally changes under independent rotations. These
are algebraic invariances of the local transition objective; the
neighboring-layer term reuses the same transition representation to compare
adjacent-layer relations.

\section{Ablation Definitions}
\label{app:ablation-defs}

Table~\ref{tab:ablations} uses short row names to keep the main paper compact.
All rows keep the OPSD output objective, the same training schedule, the same
$128$-position hidden window, and the same privileged teacher source as PHF.
Only controls evaluated at all three model scales are included in the main
ablation table.

\paragraph{What is matched.}
Direction only keeps $\lossdir$ and removes the Gram geometry term. Geometry
only keeps $\lossgeo$ and removes direction matching.

\paragraph{Layer choice.}
Selected layers uses a fixed hand-picked subset of transformer layers instead
of aggregating the transition objective over all layers.

\section{Reference-Dependence and No-Label Scope}
\label{app:reference-scope}

PHF inherits the privileged-information assumption of OPSD. The teacher forward
pass conditions on a verified reference solution, so PHF is not a no-label or
self-verifying recipe. Wrong or misaligned references can corrupt both output and
hidden-process targets, so our experiments stay within the same verified-reference
setting as OPSD and always evaluate the online student.

\section{Broader Impact}
\label{app:impact}

PHF modifies the training objective of an existing OPSD pipeline and does not
introduce new data, new models, or deployment-time capabilities. It inherits the
broader-impact profile of OPSD and knowledge distillation: better reasoning may
help downstream applications, but does not mitigate risks of the base models or
their training data.

\paragraph{Reproducibility.}
Our implementation extends the open-source OPSD codebase with hidden-state
extraction, transition normalization, direction/Gram losses, neighboring-layer
consistency, and the EMA update in \texttt{opsd\_trainer.py}; no dependency
beyond PyTorch, Transformers, and vLLM is added. Training uses the official OPSD
math set of $29{,}434$ examples, while evaluation uses the public AIME 2024,
AIME 2025, and HMMT 2025 sets under the protocol described above.
All runs use one $8{\times}$NVIDIA H20 node; the total budget for the main
results and ablations is about $600$ GPU-hours. Code and trained checkpoints
will be released upon acceptance.